\documentclass{article}

\usepackage[preprint]{neurips_2024}

\usepackage{natbib}
\setcitestyle{numbers,square}

\usepackage[utf8]{inputenc} 
\usepackage[T1]{fontenc}    
\usepackage{hyperref}       
\usepackage{url}            
\usepackage{booktabs}       
\usepackage{amsfonts}       
\usepackage{nicefrac}       
\usepackage{microtype}      
\usepackage{xcolor}         

\usepackage{graphicx}
\usepackage{float}
\usepackage{subfigure}
\usepackage{multirow}
\usepackage{makecell}
\usepackage{bbding}
\usepackage{color,xcolor,colortbl}
\usepackage[noend]{algpseudocode}
\usepackage{algorithmicx,algorithm}
\usepackage{multicol}

\definecolor{lightgray}{gray}{0.95}
\definecolor{color3}{gray}{0.95}
\definecolor{rouse}{rgb}{0.981,0.961,0.941}
\definecolor{light-yellow}{rgb}{1,1,0.93}
\definecolor{light-green}{rgb}{0.95,1,0.95}

\definecolor{colorTabTop}{rgb}{0.93,0.92,0.94} %
\definecolor{colorTab}{rgb}{0.92,0.95,0.92} %
\usepackage{microtype}

\usepackage{amsmath}
\usepackage{amssymb}
\usepackage{mathtools}
\usepackage{amsthm}

\usepackage{enumerate}
\usepackage{wrapfig}

\hypersetup{
colorlinks=true,
linkcolor=red,
}

\title{On-Chip Hardware-Aware Quantization for\\Mixed Precision Neural Networks}

\vspace{-0.3in}
\author{%
  Wei Huang\thanks{Equal contribution}\\
  The University of Hong Kong
  \And
  Haotong Qin\footnotemark[1]\\
  ETH Zürich
  \And
  Yangdong Liu\\
  Beihang University
  \And
  Jingzhuo Liang\\
  Beihang University
  \And 
  Yulun Zhang\\
  ETH Zürich
  \And
  Ying Li\thanks{Corresponding author}\\
  Beihang University
  \And
  Xianglong Liu\\
  Beihang University
}

\begin{document}

\maketitle

\vspace{-0.25in}
\begin{abstract}
\vspace{-0.1in}
Low-bit quantization emerges as one of the most promising compression approaches for deploying deep neural networks on edge devices. Mixed-precision quantization leverages a mixture of bit-widths to unleash the accuracy and efficiency potential of quantized models. However, existing mixed-precision quantization methods rely on simulations in high-performance devices to achieve accuracy and efficiency trade-offs in immense search spaces. This leads to a non-negligible gap between the estimated efficiency metrics and the actual hardware that makes quantized models far away from the optimal accuracy and efficiency, and also causes the quantization process to rely on additional high-performance devices. In this paper, we propose an \textbf{On-Chip Hardware-Aware Quantization} (OHQ) framework, performing hardware-aware mixed-precision quantization on deployed edge devices to achieve accurate and efficient computing. Specifically, for efficiency metrics, we built an \textit{On-Chip Quantization Aware} pipeline, which allows the quantization process to perceive the actual hardware efficiency of the quantization operator and avoid optimization errors caused by inaccurate simulation. For accuracy metrics, we propose \textit{Mask-Guided Quantization Estimation} technology to effectively estimate the accuracy impact of operators in the on-chip scenario, getting rid of the dependence of the quantization process on high computing power. By synthesizing insights from quantized models and hardware through linear optimization, we can obtain optimized bit-width configurations to achieve outstanding performance on accuracy and efficiency. The quantization process occurs entirely on-chip without additional devices and data access. We evaluate inference accuracy and acceleration with quantization for various architectures and compression ratios on hardware. OHQ achieves 70\% and 73\% accuracy for ResNet-18 and MobileNetV3, respectively, and can reduce latency by 15$\sim$30\% compared to INT8 on real deployment. The results show the proposed OHQ surpasses the existing state-of-the-art mixed-precision quantization methods, showcasing the significant practicality of our proposed method.
\end{abstract}

\vspace{-0.2in}
\section{Introduction}
\vspace{-0.05in}
Deep neural networks (DNNs) have developed significantly and shown great potential in various fields, such as computer vision and natural language processing. The advancement of DNNs can be mainly attributed to the rapid expansion of parameters and the increasing depth of the model. With the enhanced performance, serious challenges have been simultaneously posed by the over-parameterization of DNNs when they are deployed on resource-constrained edge devices. Edge chips impose stricter constraints on memory footprint, energy consumption, and latency~\cite{horowitz20141}. An intuitive solution to mitigate this issue is quantizing the full-precision weights and activations of DNNs to low-bit integers or binaries~\cite{gray1998quantization,han2015deep,lin2017runtime,zhu2016trained}.

\begin{wrapfigure}[19]{r}{0.6\textwidth}
  \vspace{-0.22in}
  \begin{center}
    \includegraphics[width=0.6\textwidth]{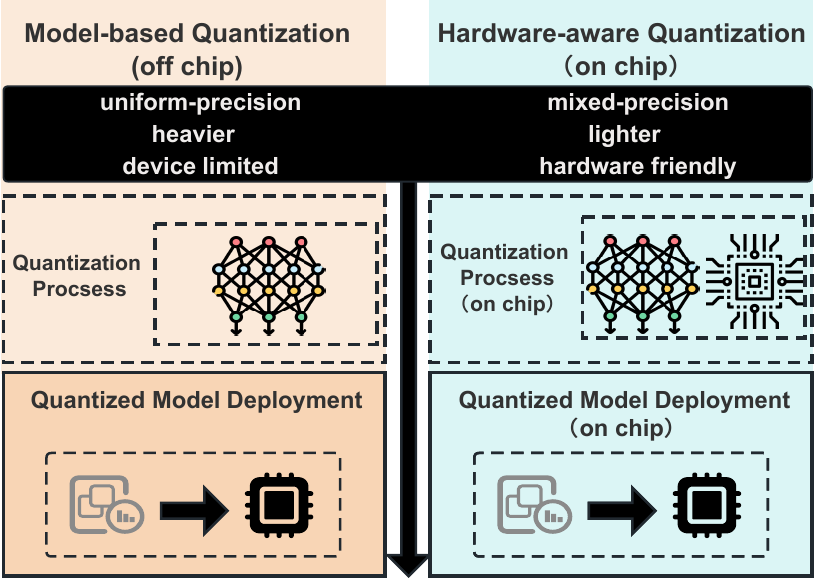}
  \end{center}
  \vspace{-0.2in}
  \caption{Off-chip \textit{vs.} on-chip quantization. The left shows the traditional off-chip quantization framework involving quantization analysis and deployment steps. The right part is our OHQ framework, which is fully integrated on-chip.}
  \vspace{-0.1in}
  \label{fig:speed-size-acc}
\end{wrapfigure}
The general quantization solution uniformly compresses a DNN to consistent bit-width~\cite{krishnamoorthi2018quantizing,choi2018pact}, allowing the model to benefit from compact low-bit parameters and integer instructions. However, this unified strategy also neglects the different sensitivities of each layer to quantization, including differences in accuracy and hardware efficiency.
To address this limitation, the mixture of bit-widths should be adopted to distinct layers, a concept known as mixed-precision quantization~\cite{cai2020zeroq,wu2018mixed,yao2021hawq,ma2023ompq}. Mixed-precision quantization considers operators with different bit widths in search space optimization, significantly pushing the accuracy and efficiency limits of quantized NNs. 
However, the existing mixed-precision quantization processes cause significant computation overhead due to the massive accuracy and efficiency metrics search space. For a $L$-layers DNN, the $S$ number of bit-width candidates results in a computation with O($S^L$) complexity. This exemplifies an exponential increase in computational demand for mixed-precision strategies, especially in deeper architectures.

These facts introduce strong motivation to construct an on-chip mixed-precision quantization framework that harmoniously integrates the quantization algorithm with real hardware. The challenges in constructing this quantization framework stem from two main aspects: (1) Hardware awareness: Since the need to consider the accuracy and speed of quantized models based on their deployed hardware, this framework should on-chip perceive various metrics of operators on real hardware, including but not limited to latency and memory usage. This framework should achieve hardware awareness for the deployed devices. (2) Lightweight process: Given the necessity to thoroughly consider efficiency-oriented hardware metrics, the expanded search space requires an efficient yet resource-lightweight algorithm for on-chip quantization processes. This ensures that the quantization process does not rely on external computational and data resources.
Despite the presence of certain hardware awareness and mixed-precision quantization techniques that emerged earlier~\cite{dong2020hawq,yao2021hawq,yang2021fracbits,ma2023ompq,cai2020zeroq,wang2019haq}, they still exhibit some significant issues that cannot be overlooked. These issues include the computationally expensive nature of the process and reliance on inaccurately estimated hardware metrics, among others. These factors hinder their progression toward becoming the ideal on-chip quantization methods.

In this work, we propose an On-Chip Hardware-Aware Quantization (OHQ) framework (see Fig.~\ref{fig:quantization flow}) to overcome the above-mentioned issues. The proposed OHA mainly relies on two novel techniques: the \textit{On-chip Quantization Awareness} (OQA) pipeline enables perceiving the actual efficiency metrics of the quantization operator on the hardware, which uses synthetic data as input to obtain the latency, memory usage, and power metrics on-chip. Second, we propose \textit{Mask-guided Quantization Estimation} (MQE) technique to efficiently estimate the accuracy metrics of operators under the constraints of on-chip-level computing power, and then we can search for optimized bit-width configurations simplified as linear programming.
Our comprehensive experiments show that OHQ outperforms existing mixed-precision quantization methods in accuracy and efficiency by a substantial margin. We also demonstrate the effectiveness of our OHA on various architectures, such as ResNet-18/50 and MobileNetV2/V3, highlighting its versatility. We summarize our contributions as follows:

(1) We propose a mixed-precision quantization framework OHQ, which is a totally on-chip design from quantization computation to deployment. By employing edge chips as co-processors alongside on-chip central processing units (CPUs) for the quantization, compression, and inference computations of models, we effectively eliminate the necessity for auxiliary computing devices. 

(2) A hardware-aware OQA that operates at the Intellectual Property (IP) core granularity is first proposed. This approach utilizes the chip's clock cycle consumption per layer as a metric for hardware while simultaneously considering the constraints imposed by the available computational power.

(3) We develop an enhanced and statistically robust sensitivity metric MQE for performing small-batch distilled data inference on edge devices, supporting a data-free on-chip pipeline of quantization.

(4) We additionally furnish comprehensive empirical findings for ResNet18/50~\cite{he2016deep}, MobileNetV2/V3~\cite{howard2019searching}. These findings delineate both the state-of-the-art quantization outcomes achievable in real-world edge scenarios, and further illustrate that this mixed-precision strategy is applicable to both the Post-Training Quantization (PTQ) and Quantization-Aware Training (QAT) approach.

\section{Related Work}

\textbf{Mixed-Precision Quantization}.
Quantization compresses $\boldsymbol{x}$ to fixed-point using the equation:
\begin{equation}
    \operatorname{quantize}(\boldsymbol{x}) = \operatorname{round}\left({\boldsymbol{x}}/{S}\right) - \boldsymbol{z},
\end{equation} 
where $\operatorname{quantize}$ denotes quantization function, $S$ and $\boldsymbol{z}$ denote the scaling factor and zero-point, respectively. 
Existing quantization approaches can be divided into quantization-aware training (QAT) and post-training quantization (PTQ). 
QAT quantizes the network throughout the training process with the original dataset, resulting in accurate quantization while markedly computation overhead~\cite{wang2019haq,yao2021hawq,choi2018pact}.
PTQ operates as an offline algorithm, it relies on a few real or synthetic samples to calibrate the quantization functions of weights and activations, thus just utilizing much less computation in the quantization process compared to QAT~\cite{cai2020zeroq,li2021brecq,nagel2019data}.

\begin{figure*}
    \centering
    \vspace{-0.5in}
    \includegraphics[width=1\linewidth ]{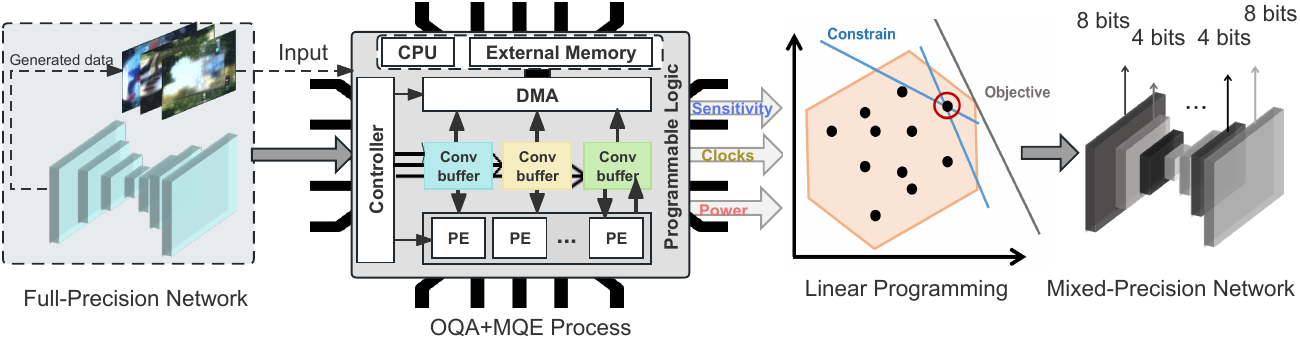}
    \vspace{-0.25in}
    \caption{The overview of OHQ framework. This proposed OHQ obtains chip-level sensing parameters and layer-wise differences through a physical deployment (OQA and MQE are respectively described in detail in Fig.~\ref{fig:Hardware_Awareness} and Fig.~\ref{fig:sensitivity}). 
    }
    \vspace{-0.2in}
    \label{fig:main flow}
\end{figure*}

For utilizing the accuracy and efficiency potential of the quantized model, mixed-precision quantization emerges as a promising way, which enables accuracy-sensitive layers to retain high precision (i.e., more bit-widths) while others maintain low precision~\cite{dong2020hawq,ma2023ompq,wang2019haq,zhou2018adaptive,shen2020q,dong2023emq}. Nevertheless, a major challenge associated with this strategy lies in identifying the optimized mixed-precision configuration since the search space exhibits an exponential relationship with the number of layers, particularly in resource-limited scenarios. Thus, various mixed-precision quantization methods are proposed to improve this. Dong et al. use a second-order Hessian matrix as the sensitivity metric for each layer~\cite{dong2019hawq} and also apply the mean of the eigenvalues of the Hessian to mitigate computational overhead~\cite{dong2020hawq}.
\cite{qin2023diverse} and \cite{cai2020zeroq} propose data-free methods to get rid of the reliance on the data resource.
However, it is still hard for existing methods to achieve accurate and efficient on-chip quantization.

\textbf{Hardware-Aware Quantization}.
The accuracy sensitivity of quantized layers can frequently be described as the influence of quantization on the model accuracy. However, the computational expenditure of network operators on physical hardware should also serve as a constraint for bit-width configuration, incorporating aspects such as latency and energy consumption. The heavily consuming layers should be quantized to low precision, while the light-consuming ones should be retained to high precision. Wang \textit{et al.}~\cite{wang2019haq} investigated the allocation of bit-width across various hardware architectures using a reinforcement learning model. This approach, described as a black-box strategy, relies on the perception of hardware characteristics. However, this exploration approach is computationally complex and difficult to realize offline quantization. Yao \textit{et al.}~\cite{yao2021hawq} identified the computation time of each network layer as a constraint for allocating bit-widths, implementing this approach by deploying the quantized network on a Graphics Processing Unit (GPU). However, the roughly obtained computation time is affected by software, operating system (OS), and network transmission, and cannot reflect the on-chip computational efficiency accurately. 

We select FPGA to implement and evaluate our framework since it serves as a flexible and reliable hardware platform for DNN deployment\cite{guo2017survey,shawahna2018fpga} and validating Application-Specific Integrated Circuit (ASIC) designs through accurate flow verification methodologies\cite{markovic2007asic,hutton2006methodology,farooq2021pre,boutros2018you}. We proposed a fine-grained hardware awareness by clock cycles and energy, which can achieve a more accurate representation of hardware computation consumption at the IP core level. This methodology enables a closer examination of the chip's underlying layer, thereby facilitating a realistic construction of the computational connection between DNNs and chips.

\section{Methodology}
In this section, we present On-Chip Hardware-Aware Quantization (OHQ) scheme (Fig.~\ref{fig:main flow}), including the On-Chip Quantization Awareness (OQA) and Mask-Guided Quantization Estimation (MQE).

\subsection{On-Chip Quantization Awareness}

We first propose an On-Chip Quantization Awareness (OQA) pipeline, which aims to enable the proposed quantization framework to perceive efficiency metrics in real-time on hardware and to get rid of the dependence on data and computation resources outside the chip.

\subsubsection{Data Preparation for Quantization}
Since the quantization process is performed on-chip, limited resources and offline scenario make hardware awareness require small batches of synthetic data.
We construct a generative data preparation that employs the statistics of BN layers, where there is the information from training data of full-precision models~\cite{nagel2019data,cai2020zeroq}. The generation process aims to facilitate the unity between the feature statistics of synthetic data $x^d$ with BN statistics:
\begin{equation}
\mathop{\min}_{x^d}\mathcal L_\text{prep} = \sum_{i=1}^L \parallel \tilde{u}_i^d - {u}_i\parallel_2^2 + \parallel \tilde{\sigma}_i^d - {\sigma}_i\parallel_2^2 \label{2},
\end{equation} 
where ${u}_i$ and ${\sigma}_i$ are the mean and standard deviation parameters of the pre-trained model's BN layer. $\tilde{u}_i^d$ and $\tilde{\sigma}_i^d$ represent the mean and standard, respectively, of the feature matrix generated by $x^d$ at the $i$-th of the BN layer. Eq.~\eqref{2} is to minimize the statistical loss $\mathcal L_\text{prep}$ in each BN layer of the original model, and ultimately to generate synthetic data that matches the input data distribution. 
The well-prepared data can be utilized for hardware awareness and quantization process, getting rid of the reliance on data out of the chip and bolstering the efficiency of the proposed framework.

\subsubsection{On-Chip Awareness of Hardware Metrics}

\begin{wrapfigure}[23]{r}{0.6\textwidth}
  \vspace{-0.3in}
  \begin{center}
    \includegraphics[width=0.6\textwidth]{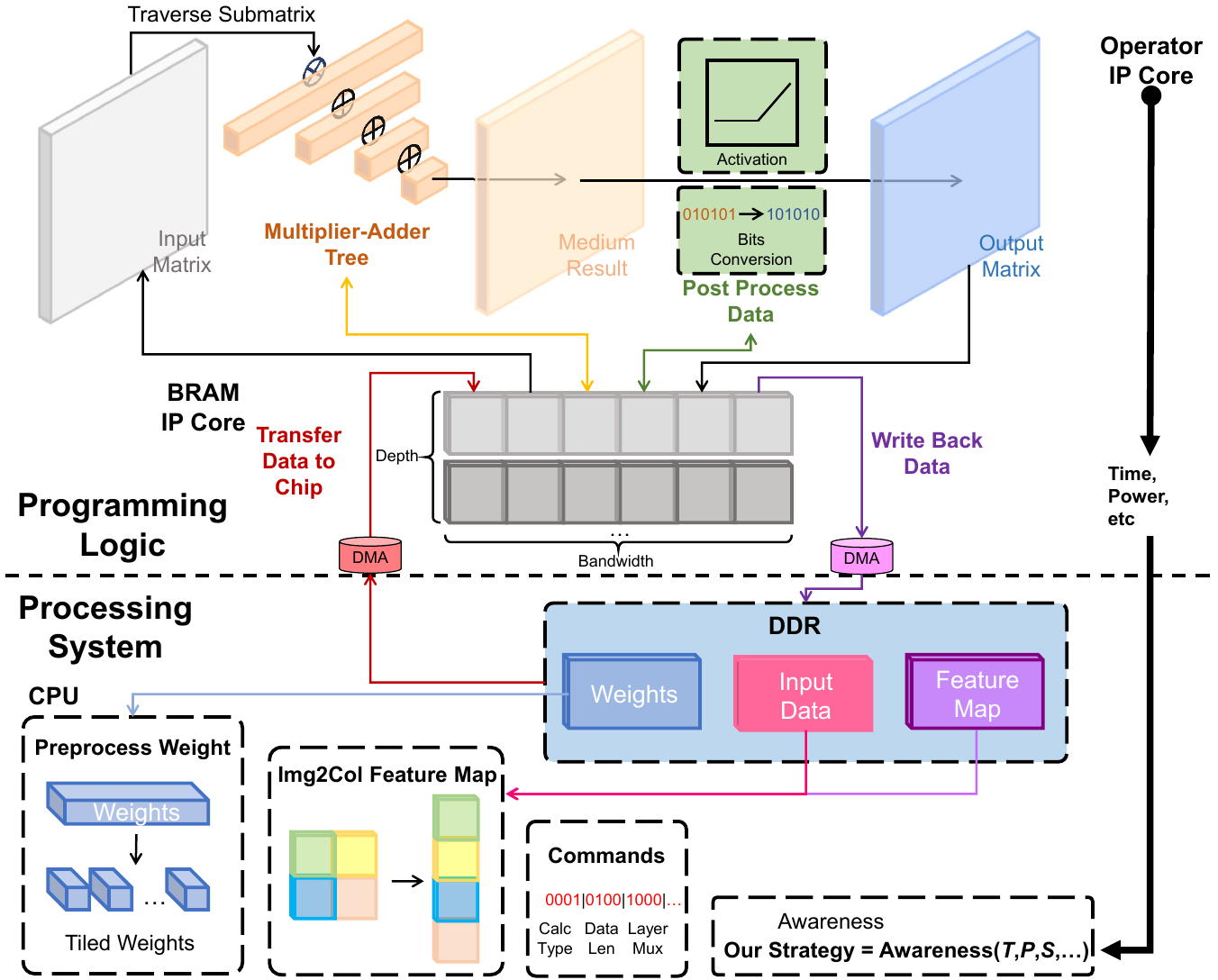}
  \end{center}
  \vspace{-0.2in}
  \caption{The workflow of OQA. (Top) The PL part samples time, power, and other information of four main steps for awareness while computing, which use BRAM to optimize matrix multiplication and data transfer. (Bottom) The PS part controls the whole situation, including accessing data, organizing the network, and instructing IP cores.}
  \vspace{-0.1in}
  \label{fig:Hardware_Awareness}
\end{wrapfigure}
Previous methods~\cite{yao2021hawq} use the running time cloud GPU to simulate the hardware metrics on edge devices, but this coarse-grained metric is influenced by the latency from operating systems and applications, thus cannot provide a comprehensive and accurate reflection on the physical chip~\cite{nurvitadhi2016accelerating,yang2012implementation}. 
To allow the quantization pipeline to be aware of the finer-grained hardware constraints accurately, we observed the IP core-level network operations on real hardware and named it OQA.
We use Field Programmable Gate Arrays (FPGAs), a flexible hardware for verification chip designs, to implement and verify our proposed method.

To enable on-chip deployment, we employ a two-step process: first, quantizing and compiling the model post-acquisition, and then translating the layers into their respective arithmetic module implementations. This is complemented by the introduction of a central controller.
1) Img2Col: converts layer data from discontinuous storage to continuous storage, streamlining transmission and computation.
2) Multiplication and addition tree: utilizes parallel stacking and cascading multiplication and addition mechanisms to enable simultaneous pointwise multiplication computation across vast amounts of data.
3) Sub-matrix slicing transmission and computation: divides a large matrix into multiple tiles for transmission to the chip, performing computation and splicing on these tiles to alleviate FPGA resource pressure.
4) BRAM large bandwidth fill: the primary storage mechanism employed in on-chip Block Random Access Memory (BRAM), which is divided into rows to facilitate the reading and writing of large bit-width data simultaneously, ensuring computation and access remain logically and physically coherent.
The above methods enable efficient parallel computation of networks, and a quantitative deployment scheme can be formulated with full consideration of the hardware resource constraints. 

Our proposed clock-aware approach, predicated on the interaction between the IP core and the BRAM, is illustrated in Fig.~\ref{fig:Hardware_Awareness}. During runtime, the IP core on the Programmable Logic (PL) side—which is responsible for four steps including computation of weight and feature map, data transfer, data write-back, and data post-process—automatically collects the number of running clock cycles and stores them in the target BRAM. 
Then, the transmission mechanism relays the collected clock cycles to the Processing System (PS) side. This can be denoted as $[c_1,\dots,c_i,\dots,c_L]$, $c_i$ represents the total number of clock cycles in the $i$-th layer. Concurrently, the power consumption of the IP core during computation is diminished as $[e_1,\dots,e_i,\dots,e_L]$.

Note that OQA is the first pipeline to propose fine-grained sensing at the chip IP core level. By obtaining the clock information of each of the four calculation steps from the PL, we get the exact part of the operation of each layer on-chip to optimize the following MQE. 

\subsection{Mask-Guided Quantization Estimation}
We propose Mask-Guided Quantization Estimation (MQE) to determine mixed-precision configurations for quantization with comprehensive efficiency and accuracy.

\subsubsection{On-Chip Awareness of Layerwise Sensitivity}
\begin{wrapfigure}[12]{r}{0.7\textwidth}
  \vspace{-0.3in}
  \begin{center}
    \includegraphics[width=0.7\textwidth]{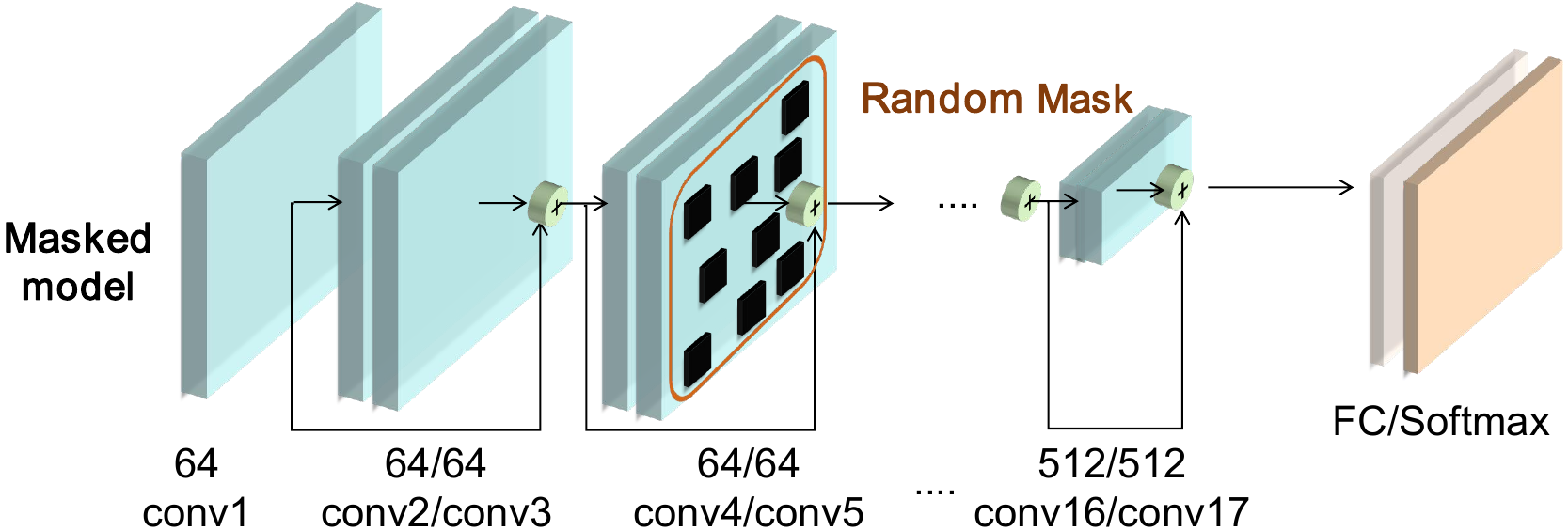}
  \end{center}
  \vspace{-0.15in}
  \caption{Illustration of MQE for ResNet18. Specifically, we feed synthesized data into on-chip models. The figure shows the model with the $5$-th layer specifically masked out.}
  \vspace{-0.1in}
  \label{fig:sensitivity}
\end{wrapfigure}
DNN models comprise $L$ layers of computational units, which can be represented as $[U_1, U_2,..., U_L]$, wherein $U_i$ denotes the $i$-th layer computational unit. The learnable parameters (e.g., weights) are denoted as $[\theta_1,\theta_2,...,\theta_L]$, where $\theta \in \mathcal R^n$ is float32 type data in full-precision models. 
Layer-wise sensitivity typically corresponds to the impact of distinct layers within a network on the output results. As previously discussed, sensitivity measurement using the Hessian matrix can be computationally intensive. An alternative approach involves quantizing  $U_i$ to 4/8 bits while maintaining full precision for the remaining layers:
\begin{equation}
    {\theta_i^q} = \operatorname{quantize}({\theta_i}),
\end{equation}
\begin{align}
\left\{\begin{aligned}
&\mathcal M =   F(U_1(\theta_1);\dots;U_i(\theta_i);\dots;U_L(\theta_L)),\\
&\mathcal M_i^q =   F(U_1(\theta_1);\dots;U_i^q(\theta_i^q);\dots;U_L(\theta_L)),
\end{aligned}\right.
\end{align}
where $\theta_i^q$ is the quantized parameters in $i$-th layer and the bit width, $q$ is the selected bit-widths, $F$ denotes the general function of NN model, $\mathcal M$ represents the full-precision model, $\mathcal M_i^q$ denotes the quantized model with quantized layer $U_i^q$. Then, the performance difference is calculated by the following equation:
\begin{equation}
    \omega_i = \frac{1}{N}\sum_{j=1}^N {f(\mathcal M(x_j), \mathcal M_i^q(x_j))},\label{5}
\end{equation}
where $\omega_i$ indicated the sensitivity value of the $i$-th layer, $N$ is the batch-size of distilled data used for inference, $f(\cdot , \cdot)$ denotes the sensitivity calculation function that compares the output of $\mathcal M$ and $\mathcal M_i^q$, and $x$ is the input data. 

Nonetheless, this method necessitates the quantization $\textbf{L}$ times, resulting in the considerable computational overhead of the quantization process. 
Consequently, this sensitivity calculation presents challenges for on-chip deployment. Therefore, we introduce a masked-guided technique (MQE) displayed in Fig.~\ref{fig:sensitivity}, which is more efficient and suitable for on-chip implementation:
\begin{equation}
    \tilde{\theta}_i^q = g(\alpha, \theta_i^q),
\end{equation}
\begin{align}
\left\{\begin{aligned}
&\mathcal M^{q} =  \ F(U_1^q(\theta_1^q);\dots;U_i^q(\theta_i^q);\dots;U_L^q(\theta_L^q)),\\
&\mathcal {\tilde{M}}_i^q = \ F(U_1^q(\theta_1^q);\dots;\tilde{U}_i^q(\tilde{\theta}_i^q);\dots;U_L^q(\theta_L^q)),
\end{aligned}\right.
\end{align}
where $g(\cdot,\cdot)$ denotes the masking operator, $\alpha$ is the mask ratio, and $\tilde{\theta}_i^q$ is the mask result of $\theta_i^q$. To facilitate the on-chip test of FPGA, the parameters must first undergo integer bit-wise quantization (in this case, $q = 8$). Subsequently, the parameters of the target layer are randomly masked by setting them to 0, and the $\alpha$ is selected as $0.5$ to map the loss of information loss from $8$ bits to $4$ bits.  In MQE, only \textbf{1} time quantization of the DNN is necessary. Despite the requirement for $L$ masking operations, their computational consumption is considerably less compared to that of quantization computation. Then, we use the KL divergence:
\begin{equation}
D_{kl}(p \parallel q) = -\sum_x p(x)\log q(x) - \sum_x -p(x)\log p(x),
= \sum_x p(x)\log{\frac{p(x)}{q(x)}},\label{8}
\end{equation}
where $p(x)$ and $q(x)$ denote two probability distributions of the original model and masked model. KL divergence can determine the disparity in output distribution between the original and masked models, thereby effectively measuring the information entropy between the matrices. Then, we update Eq.~\ref{5} to get our sensitivity:
\begin{equation}
    \omega_i= \frac{1}{N}\sum_{j=1}^N {D_{kl}(\mathcal M^{q}, \mathcal {\tilde{M}}_i^q)},
\end{equation}
Our proposed MQE is obtained by inference on edge devices, which better reflects the layer-wise perception in NN in real hardware operation scenarios than simulation experiments on servers and GPUs. Consequently, the masking operations are negligible and can be swiftly executed by the CPU embedded within the edge device.

\subsubsection{Mixed-Precision Quantization Strategy}





Our findings reveal that the relationship between hardware consumption and model parameters is not entirely linear, demonstrating variability as depicted in Fig.~\ref{ablition_exp}. This observation validates the efficacy of our proposed OHQ framework in precisely assessing the performance of both the network and the hardware, underscoring the critical role of on-chip hardware awareness in the quantization process. Motivated by these insights, we aim to strike an optimal balance between model performance and hardware computational efficiency. Towards this end, we have formulated a quantitative constraint function that amalgamates On-chip OQA and MQE:
\begin{equation}
    1 = \beta + \gamma,\qquad
    \Omega_i = \beta \hat{\omega}_i - \frac{\gamma}{2}(\hat{c}_i + \hat{e}_i),
    \label{eq11}
\end{equation}
where $\beta$ and $\gamma$ are two hyper-parameters used to control the proportion of the sensitivity and hardware resources. To consider this hardware awareness fairly, we set $\beta$ and $\gamma$ to 0.5 in the subsequent experiments. The hyper-parameters can be manually modified to satisfy the personalized deployment for the precision and hardware compression rate. $\hat{\omega}_i$, $\hat{c}_i$ and $\hat{e}_i$ are scaled values from $\omega_i$, $c_i$ and $e_i$, which ensures that the different initial awareness value is in same the range $[0,1]$ and $\Omega_i$ is the optimal factor of the $i$-th layer. Ultimately, we maximize the sum of  $\Omega_i$ in the network through an integer linear programming (ILP) model:
\begin{align}
\text{Objective}:\max_{\{b_i\}^L_{i=1}} \sum^L_{i=1}(b_i \Omega_i),\quad
\text{Constrain}:\sum^L_{i=1} M_i^{b_i} \leq \text{Model\ Size\ Limit},
\end{align}
$M_i^{b_i}$ denotes the parameters size of the $i$-th layer under $b_i$ bit-width quantization. The compression ratio of the target model is flexible . Since the selectable bit widths are 4 and 8, the target size should be between the uniform 4-bit model and the uniform 8-bit model. It is worth noting that compared to the reinforcement learning method proposed by Wang et al.\cite{wang2019haq} and the Hessian evaluation proposed by Dong et al.\cite{dong2020hawq}, the ILP model only takes about 1 second to obtain the optimal bit-width configuration result under the hardware-aware result input condition. This computation is very efficient and can be done on the embedded CPU of the edge platform. 


\begin{table*}[t]
\vspace{-0.2in}
\small
\centering
\setlength{\tabcolsep}{1.mm}
\begin{tabular}{llcccrrrr}
\hline
\rowcolor{colorTabTop}
Arch&Method & Int-Only & Uniform & W/A & Data & Size (Mb) & Latency(ms)&Top-1 (\%) \\

\hline\hline


&Baseline & \XSolid & - & 32/32 & 1.2E6 & 44.6&39.6$^\star$& 73.21 \\
\cline{2-9}
&Min\&Max & \XSolid & \Checkmark & 8/8&1.2E6&11.1&78.3&71.38\\
ResNet18&\cellcolor{colorTab}\textbf{OHQ (ours)}& \cellcolor{colorTab}\Checkmark & \cellcolor{colorTab}\Checkmark & \cellcolor{colorTab}8/8&\cellcolor{colorTab}$^\dag$\textbf{32} &\cellcolor{colorTab}\textbf{11.1}&\cellcolor{colorTab}\textbf{78.3}&\cellcolor{colorTab}\textbf{71.52}\\ 
\cline{2-9}
&FracBits-PACT\ddag\cite{choi2018pact}& \XSolid & \Checkmark &*/*&1024&5.8&68.6 &69.70 \\
&ZeroQ\cite{cai2020zeroq}& \Checkmark & \Checkmark & */*& $^\dag$32& 5.8&67.9 &21.20 \\
&BRECQ\cite{li2021brecq}& \Checkmark & \Checkmark &*/8&1024&5.8&67.6&69.32 \\
&OMPQ\cite{ma2023ompq}& \Checkmark & \Checkmark &*/4&64&5.5&70.2&69.38 \\
&EMQ\cite{dong2023emq}& \Checkmark & \Checkmark &*/4&64&5.5&- &70.12 \\

&\cellcolor{colorTab}\textbf{OHQ (ours)}& \cellcolor{colorTab}\Checkmark & \cellcolor{colorTab}\Checkmark &\cellcolor{colorTab}*/*&\cellcolor{colorTab}$^\dag$\textbf{32}\cellcolor{colorTab}&\cellcolor{colorTab}\textbf{5.5}&\cellcolor{colorTab}\textbf{63.5}&\cellcolor{colorTab}\textbf{70.18} \\

\hline


&Baseline & \XSolid & - & 32/32 & 1.2E6 & 97.8 &80.2$^\star$& 77.72 \\
\cline{2-9}
&Min\&Max & \XSolid & \Checkmark & 8/8&1.2E6&24.5&182.6&{77.70}\\
ResNet50&\cellcolor{colorTab}\textbf{OHQ (ours)}& \cellcolor{colorTab}\Checkmark & \cellcolor{colorTab}\Checkmark & \cellcolor{colorTab}8/8&\cellcolor{colorTab}$^\dag$\textbf{32}&\cellcolor{colorTab}\textbf{24.5}&\cellcolor{colorTab}\textbf{182.6}&\cellcolor{colorTab}\textbf{77.72}\\ 
\cline{2-9}
&PACT\ddag\cite{choi2018pact}& \XSolid & \Checkmark &*/*&1024&19.2&151.2 &75.3 \\
&OCS\cite{zhao2019improving}& \Checkmark & \Checkmark & 6/6& 1.2E6& 18.4 &159.6& 74.80\\
&FIOPTQ\cite{chauhan2023post}& \Checkmark & \Checkmark & */8 & 2048 & 12.2 & - & 75.44\\
&ZeroQ\cite{cai2020zeroq}& \Checkmark & \Checkmark & */6& $^\dag$32 & 18.3 &160.3&77.43\\

&\cellcolor{colorTab}\textbf{OHQ (ours)}& \cellcolor{colorTab}\Checkmark & \cellcolor{colorTab}\Checkmark &\cellcolor{colorTab}*/*&\cellcolor{colorTab}$^\dag$\textbf{32}&\cellcolor{colorTab}\textbf{17.8}&\cellcolor{colorTab}\textbf{147.8}& \cellcolor{colorTab}\textbf{77.55} \\

\hline


&Baseline & \XSolid & - & 32/32 & 1.2E6 & 13.4 &11.3$^\star$& 73.03 \\
\cline{2-9}
&Min\&Max & \XSolid & \Checkmark & 8/8&1.2E6&3.3&100.7&70.29\\

MobileNetV2&\cellcolor{colorTab}\textbf{OHQ (ours)}& \cellcolor{colorTab}\Checkmark & \cellcolor{colorTab}\Checkmark & \cellcolor{colorTab}8/8&\cellcolor{colorTab}$^\dag$\textbf{32} &\cellcolor{colorTab}\textbf{3.3}&\cellcolor{colorTab}\textbf{100.7}&\cellcolor{colorTab}\textbf{73.00}\\ 
\cline{2-9}
&FIOPTQ\cite{chauhan2023post}& \Checkmark & \Checkmark & */16 & 2048 & 3.4 & - & 71.38\\
&FracBits-PACT\ddag\cite{choi2018pact}& \Checkmark & \Checkmark & */*& $^\dag$32&1.8&85.4&69.90\\
&BRECQ\cite{li2021brecq}& \Checkmark & \Checkmark &*/8&1024&1.5&81.7 &70.28 \\
&EMQ\cite{dong2023emq}& \Checkmark & \Checkmark &*/8&64&1.5&- &70.75 \\
&\cellcolor{colorTab}\textbf{OHQ (ours)}& \cellcolor{colorTab}\Checkmark & \cellcolor{colorTab}\Checkmark &\cellcolor{colorTab}*/*&\cellcolor{colorTab}$^\dag$\textbf{32} &\cellcolor{colorTab}\textbf{1.5}&\cellcolor{colorTab}\textbf{67.1}&\cellcolor{colorTab}\textbf{71.46} \\

\hline


&Baseline & \XSolid & - & 32/32 & 1.2E6 & 15.3 &10.6$^\star$&74.32 \\
\cline{2-9}
&Min\&Max & \XSolid & \Checkmark & 8/8&1.2E6&3.9&85.0&72.98\\

MobileNetV3&\cellcolor{colorTab}\textbf{OHQ (ours)}& \cellcolor{colorTab}\Checkmark & \cellcolor{colorTab}\Checkmark &\cellcolor{colorTab} 8/8&\cellcolor{colorTab}$^\dag$\textbf{32}&\cellcolor{colorTab}{\textbf{3.9}}&\cellcolor{colorTab}{\textbf{85.0}}&\cellcolor{colorTab}\textbf{74.29}\\ 
\cline{2-9}

&\cellcolor{colorTab}\textbf{OHQ (ours)}& \cellcolor{colorTab}\Checkmark &\cellcolor{colorTab} \Checkmark &\cellcolor{colorTab}*/*&\cellcolor{colorTab}$^\dag$\textbf{32}&\cellcolor{colorTab}\textbf{2.4}&\cellcolor{colorTab}\textbf{73.4}&\cellcolor{colorTab}{\textbf{73.01}} \\

\hline

\end{tabular}

\vspace{-0.1in}
\caption{
Results of PTQ. $^\dag$~indicates using distilled data in the quantization process. $^\star$~means the latency results are tested on the CPU, others are deployed on FPGA (batch=1). - is the lack of hardware deployment latency testing, which may be due to that the method has not disclosed the code or the precision preservation method is not being supported on our selected hardware chip.
}
\vspace{-0.2in}
\label{results}
\end{table*}
 
\section{Experiment} \label{exp}
In this subsection, a comprehensive array of experiments is undertaken to empirically validate the performance of the OHQ. The preliminary step encompasses the delineation of the datasets employed and the specific models selected for the experimental evaluations. The ablation experiments compare the decomposition of OQA and MQE. Conclusively, an intricate comparison is executed, juxtaposing the performance metrics of disparate models as facilitated by the OHQ framework. 
The results of this comparative assessment conspicuously highlight the discernible merits inherent in our proposed approach, manifesting in superior compression rates, heightened operational performance, and enhanced quantization efficiency.

\subsection{Implementation Details}
We validate the experiments on ImageNet\cite{deng2009imagenet} dataset. The training data (1.5M) encompassing Imagenet was deliberately left unutilized in our study. Instead, only 32 distilled images are generated, specifically dedicated to the computational assessment of perceptual outcomes about OQA and MQE during the forward inference. Owing to the wide use of residual structures in DNNS, we chose the ResNet-18 and -50. In addition, we also conduct experiments on MobileNet-V2 and -V3.

Our experimentation and assessments are exclusively executed on an ECE-EMBD development board, housing components from the ZYNQ chip series. This development board amalgamates a dual-core ARM Cortex-A9 processor endowed with 512MB of DDR3 memory on the PS side. The memory subsystem is defined by the MT41K256M16TW model, boasting a 16-bit bit width. The PL domain encompasses the XC7Z020-CLG400-1 chip, characterized by an assembly of 85K logic resources, and 140 BRAMs with a cumulative capacity of 4.9 Mb. Notably, the on-chip hardware sensing leverages a quantization strategy spanning from 4 bits to 8 bits, judiciously bounding matrix parallel operations to a maximum size of 128 data elements on a single facet. This prudent limitation is designed to preempt an overallocation of BRAM resources and thus preserve adherence to quotas. 

\subsection{Comparison Results}
The important feature of the OHQ proposed in this paper is the on-chip, and in order to efficiently implement the model, we performed PTQ on ResNet18/50, MobileNetV2, and MobileNetV3. and compared it with previous methods~\cite{cai2020zeroq,yang2021fracbits,li2021brecq,nagel2019data,zhao2019improving,park2020profit}. Table~\ref{results} underscores OHQ as the optimal 8-bit quantization strategy across diverse networks. The optimal trade-off between precision and compression ratio in mixed-precision quantization is also shown.  In the performance of ResNet18, an accuracy of 70.08\% is attained, harmoniously juxtaposed with a compact footprint of 5.8 Mb and latency of 63.5 ms. Meanwhile, the ResNet50 model, compressed to a size of 17.8 Mb with a speed of 147.8 ms, exhibits an accuracy of 77.55\%. The highest accuracy and lowest compression Rate and time were achieved in MobileNetV2 (71.64\%, 1.7 Mb, 70.1 ms). Notably, MobileNetV3 attains an accuracy of 73.01\% while preserving a remarkably small size of 2.4 Mb and latency of 73.4 ms.  

\begin{wraptable}[15]{r}{0.5\textwidth}
\vspace{-0.3in}
\caption{Results of QAT. \ddag \ means not quantizing the first and last layers. - means the lack of hardware deployment latency testing.
}
\vspace{-0.1in}
\label{results_qat}
\setlength{\tabcolsep}{0.15mm}{
\begin{tabular}{llcccc}\\\toprule  
\hline
\rowcolor{colorTabTop}
Arch&Method  & W/A  & \makecell{Size \\(Mb)} &\makecell{Latency\\(ms)}& \makecell{Top-1 \\(\%)} \\
\hline
&PACT\ddag\cite{choi2018pact} &5/5 &7.2 &70.4&69.80 \\
ResNet18&HAWQ-v3\cite{yao2021hawq} & */*&7.3 &72.9&70.01 \\
&\cellcolor{colorTab}\textbf{OHQ (ours)} &\cellcolor{colorTab}*/*&\cellcolor{colorTab}\textbf{6.9}&\cellcolor{colorTab}\textbf{68.3}& \cellcolor{colorTab}\textbf{71.87} \\
\hline
&HAWQ-v3\cite{yao2021hawq} & */*& 18.7 &172.1& 75.39 \\
ResNet50&OMPQ\cite{ma2023ompq}& */5&18.7&185.4&76.28 \\
&\cellcolor{colorTab}\textbf{OHQ (ours)} &\cellcolor{colorTab}*/*&\cellcolor{colorTab}\textbf{16.5}&\cellcolor{colorTab}\textbf{135.9}&\cellcolor{colorTab} \textbf{76.64} \\
\hline
&GZNQ\cite{he2021generative} &6/6&2.5&81.7& 71.10 \\
{\scriptsize MobileNetV2}&HAQ\cite{wang2019haq} &*/*&1.8&75.5& 71.47 \\
&\cellcolor{colorTab}\textbf{OHQ (ours)} &\cellcolor{colorTab}*/*&\cellcolor{colorTab}\textbf{1.7}&\cellcolor{colorTab}\textbf{70.1}&\cellcolor{colorTab}\textbf{72.56} \\
\hline
\end{tabular}}
\end{wraptable} 
We also combine QAT and OHQ due to the ability of the QAT method to improve quantization model accuracy through data retraining (shown in Table~\ref{results_qat}). In the ResNet18 network, we achieve, with only 6.9Mb of the model, 70.23\% accuracy, which is ahead of HAWQ-v3 in terms of compression and performance. Notably, PACT does not quantize the input and output layers, and the activations are retained 23 bits from HAQ.
For ResNet50, OHQ shows the best results(76.64\%, 16.5 Mb, 135.9 ms). Although OHQ is not specifically designed for QAT and is slightly lower in accuracy than current advanced methods, thanks to its on-chip awareness capabilities, the computing delay deployed on edge chips is minimized. Moreover, in MobileNetV3, the accuracy of OHQ is 0.08\% and 0.46\% higher than that of HAQ and GZNQ, respectively, and the compression rate of the model is 20\% higher than that of GZNQ, which still shows the sota quantization performance. Notably, all the OHQ quantization models have the fastest inference on FPGA.

\subsection{Ablation Results}
\subsubsection{Hardware-aware Parameters}
\begin{figure*}[t]
\centering  
\includegraphics[width=1.0\linewidth]{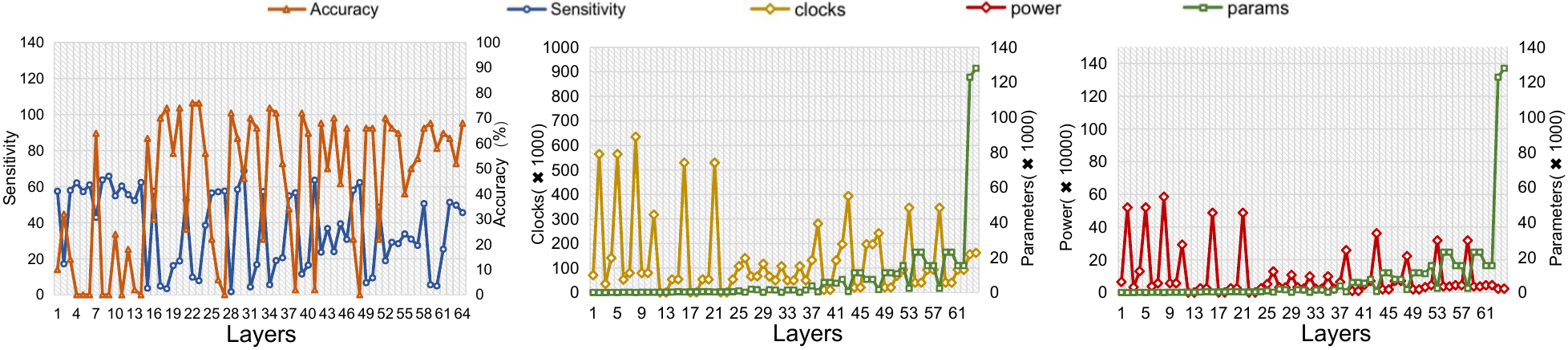}
\vspace{-0.1in}
\caption{Comparison of MobileNetV3's on-chip awareness characteristics and network parameters of each layer. The first image on the left demonstrates the relationship between the model accuracy (in \textcolor[RGB]{197,90,17}{orange}) post-application of layer-wise masking and the sensitivity of that layer (in \textcolor[RGB]{47,85,151}{blue}). The central image depicts the relationship between the on-chip computational clock (in \textcolor[RGB]{191,144,0}{yellow}) for each layer and the number of parameters in the layer (in \textcolor[RGB]{84,130,53}{green}). The image on the right presents the relationship between the on-chip power consumption (in \textcolor{red}{red}) and the number of parameters.} 
\label{ablition_exp}  
\vspace{-0.1in}
\end{figure*}

We present an analytical exposition of statistical insights in Fig.~\ref{ablition_exp} for MobileNetV3. We unveil the sensitivities exhibited across layers(left). Notably, higher sensitivities exert a profound influence on accuracy, underscoring their pivotal role in shaping model performance. We found that the first and the last layers give higher sensitivities in each network which is since the first layer directly deals with original images or feature maps with larger aspects, while the last layer needs to perform the hidden layer classification computation that determines the output data, and is prone to accumulating errors and is sensitive to the weight changes in on-chip low-precision quantitative inference scenarios. The middle demonstrates the change of clocks, and it can be seen that the latency on-chip is not strongly correlated with the parameter number. It is more strongly correlated with the size of the input feature map of each layer, while depthwise convolution has a greater correlation between the computation time and the number of channels compared to ordinary convolution. Power consumption is predominantly shaped by parallel computation scales, indicating the reduced rate of consumption increases with large size considering power and computation efficiency. More models' experiments are shown in Appendix \ref{ap_fig}.

\begin{table*}[t]
\small
\centering
\vspace{-0.4in}
\setlength{\tabcolsep}{0.3mm}
\begin{tabular}{lcccccccccc}
\hline
\rowcolor{colorTabTop}
& & & \multicolumn{2}{c}{Vanilla}& \multicolumn{2}{c}{MQE} & \multicolumn{2}{c}{OQA} & \multicolumn{2}{c}{MQE + OQA} \\

\cline{4-5} \cline{6-7} \cline{8-9} \cline{10-11}
\rowcolor{colorTabTop}
\multirow{-2}{*}{Arch} & \multirow{-2}{*}{Ratio} & \multirow{-2}{*}{W/A} &\makecell{Size (Mb)}&\makecell{Top-1 (\%)}&\makecell{Size (Mb)}&\makecell{Top-1 (\%)}&\makecell{Size (Mb)}&\makecell{Top-1 (\%)}&\makecell{Size (Mb)}&\makecell{Top-1 (\%)}  \\

\hline\hline
ResNet18

&0.25&  */*& 7.3&   70.01	&6.90&	70.23&	6.20&	69.97&	6.90&	70.23\\

&0.50&	*/*&7.9&   70.50&   8.30&     71.31&	 7.30&	70.05&	 7.70&	 71.25\\

&0.75&	*/*&9.9&  71.20&	9.70&	 71.79&	 8.20&	70.13&	 8.90&	 71.56\\

&0.80&	*/*&10.7&	 71.91&	 10.00&	 72.53&	 10.00&	71.12&	 {10.00}&	 {72.53}\\
\hline

ResNet50

&0.25&  */*& 16&   74.89	&15.47&	 {75.00}&	 {14.21}&	72.64&	 15.1&	 74.72\\

&0.50&	*/*&19&   75.95&   17.51&     76.35&	 {16.76}&	74.84&	 17.51&	 {76.35}\\

&0.75&	*/*&21.3&  77.38&	20.5&	 {77.44}&	 {19.2}&	75.01&	 19.82&	 77.19\\

&0.80&	*/*&22&	77.4&	 21.09&	 77.61&	 19.65&	76.28&	 {21.09}&	 {77.61}\\

\hline
MobileNetV2

&0.25&	*/*&2.29&	 71.22&  2.06&	 {71.62}&	 {1.97}&	70.05&	 2.06&	 {71.62}\\
&0.50&	*/*&2.51&	71.46&	2.54&	 {71.75}&	 {2.13}&	71.33&	 {2.43}&	 71.70\\
&0.75&	*/*&2.60&	 71.81&	 2.60&	 {71.83}&	 {2.36}&	71.49&	 2.60&	 {71.83}\\
&0.80&	*/*&2.75&	 72.08&	 2.70&	 {72.79}&	 {2.61}&	71.81&	 2.70&	 {72.79}\\
\hline\hline

\end{tabular}

\vspace{-0.1in}
\caption{Ablation experiments of different optimization factors on ResNet18 and MobileNetV2. We select HAWQ-v3 as our Vanilla baseline, which is a Hessian-based method. W/A is the bit-width of weight and activation. * represents mixed precision (4/8 bits).
}
\label{ablition}
\vspace{-0.1in}
\end{table*}

\subsubsection{Optimization Factor Deconstruction}

The study conducts a comprehensive examination of compression and performance dynamics across various models. HAWQ-v3 is chosen as the foundational method, which also implements mixed-precision quantization with an awareness of hardware capabilities. The results, presented in Table~\ref{ablition}, highlight the superior performance gains facilitated by MQE-based sensitivity. While OQA-guided hardware-aware constraints exhibit an appreciable augmentation in compression rates, the adoption of OQA as a sole optimization factor yields elevated accuracy loss in bit-width configurations. As demonstrated by Eq.~\ref{eq11}, integrating MQE and OQA constraints produces a synergistic effect, achieving a higher compression rate while maintaining accuracy, in sharp contrast to the results obtained from using individual parameters alone. This detailed analysis highlights the complex interrelation between compression, performance, and model constraints, aiding in the making of well-informed decisions about optimization strategies and trade-offs in resource-limited scenarios.

\textbf{Limitation}. Though with significantly improved performance, the accuracy of networks quantized by OHQ is not yet on par with their full-precision counterparts, especially under low compression ratios.

\section{Conclusion}
In this paper, our proposed OHQ introduces an innovative and effective solution for hardware-aware mixed-precision quantization, offering a substantial stride toward efficient and accurate deployment of DNNs on resource-constrained chips. Firstly, the OQA pipeline furnishes an avenue to comprehend the true efficiency metrics of quantization operators within the hardware ecosystem and yields insights that inform subsequent optimization steps. Secondly, the MQE technique is meticulously designed to efficiently gauge accuracy metrics for operators while adhering to on-chip computational constraints. Synthesizing network and hardware insights through linear programming, we derive optimal bit-width configurations. A remarkable facet of our approach is the entire quantization unfolds on-chip.


{
\small
\bibliography{arxiv}
\bibliographystyle{plain}
}

\newpage
\appendix

\section{Broader Impacts and Experiments Reproducibility}

\subsection{Broader Impacts}
In this paper, we introduce an on-chip mixed-precision technique to achieve accurate and efficient low-bit weight quantization for neural networks. This approach makes the application of neural networks more efficient and accessible, potentially broadening their widespread impact. From a positive perspective, on-chip quantization facilitates the use of neural networks. It reduces the cost and hardware barriers to deploying multitask neural networks and promotes edge inference with these models. Additionally, edge neural network computing can effectively protect user privacy and prevent data breaches. On the other hand, the generalization of AI technology may require a more comprehensive regulatory system to control its use in human society, and quantization compression technology does not contravene this principle.
\subsection{Experiments Reproducibility}
Our code is included in the supplementary materials. For instructions on how to reproduce various experiments, please refer to the code scripts and algorithm descriptions accompanying our paper. We also provide detailed information on downloading and using the datasets for the aforementioned experiments.

\section{Appendix}

\subsection{On-chip Comparison} \label{ap_fig}
\begin{figure*}[!ht]
	\centering
	\subfigure[Sensitivity]{
		\centering
		\includegraphics[width=1\linewidth]{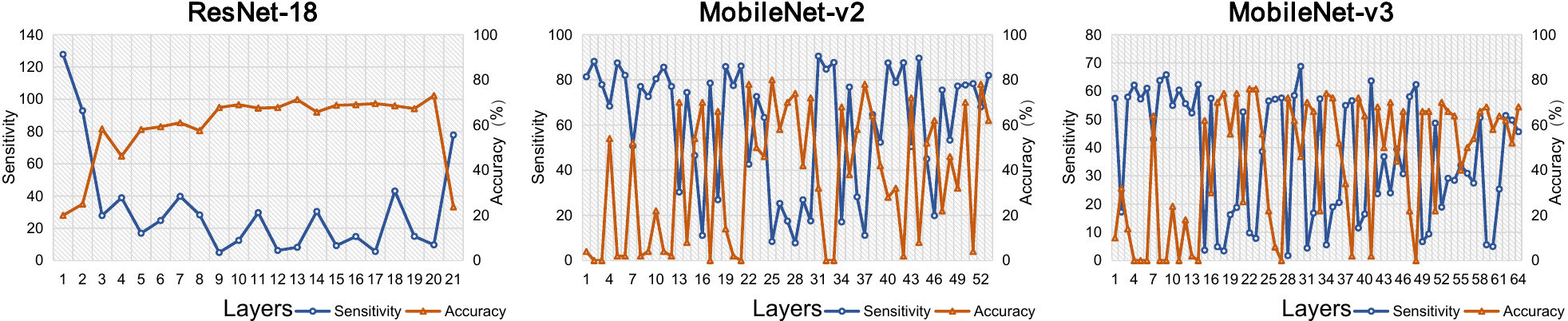}
		\label{mqe_acc}
	}
	\centering
	\subfigure[Clock]{
		\centering
		\includegraphics[width=1\linewidth]{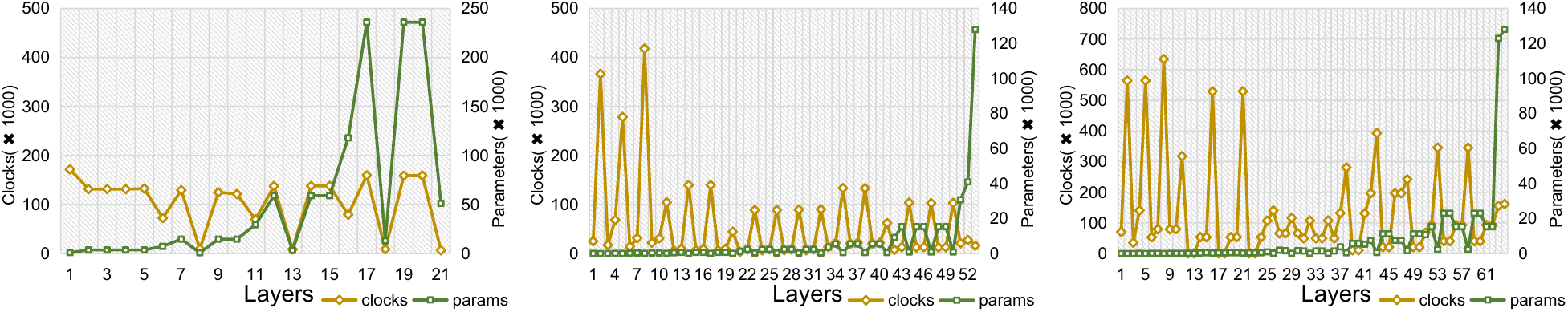}
            
		\label{oqa_clock}
	}
	\centering
	\subfigure[Power]{
		\centering
		\includegraphics[width=1\linewidth]{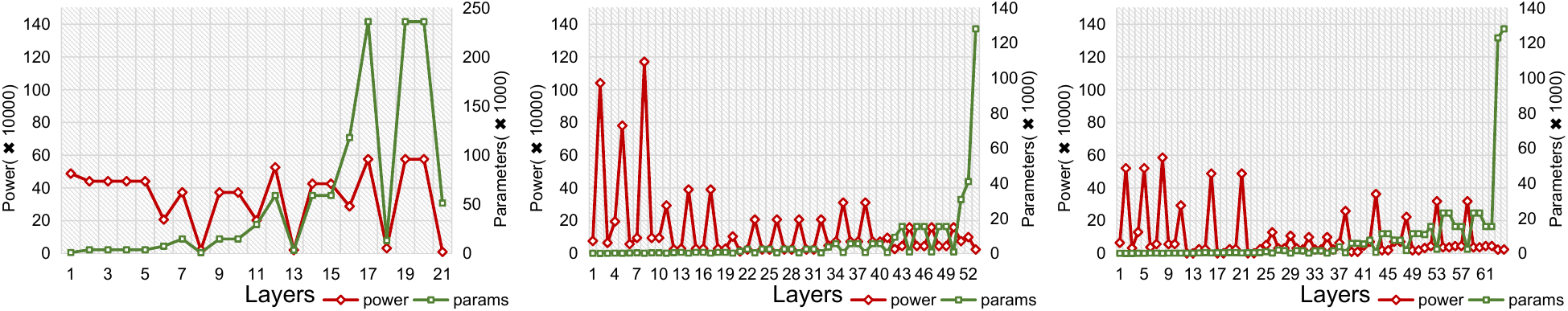}

		\label{oqa_power}
	}
	\caption{Comparison of on-chip perceptual parameters with the characteristics of the models. ResNet18, MobileNetV2, and MobileNetV3 are selected (from left to right), and the sensitivities from MQE and the clock information obtained from OQA, as well as the power consumption estimates, are obtained through the OHQ. }
	\label{ablation_all}
 
\end{figure*}

Fig.~\ref{ablation_all} further shows on-chip awareness characteristics of ResNet-18, MobileNet-v2 and MobileNet-v3. It further proves that the computing clock cycle and power consumption of the hardware cannot simply correspond to the amount of model parameters. However, OHQ obtains on-chip attributes of the neural networks more accurately through hardware perception.
\subsection{Hardware Implementation Details}
\label{impl_details}
\subsubsection{Img2Col}
\label{img2col_sec}
\begin{figure}
    \centering
    \includegraphics[width=.9\linewidth]{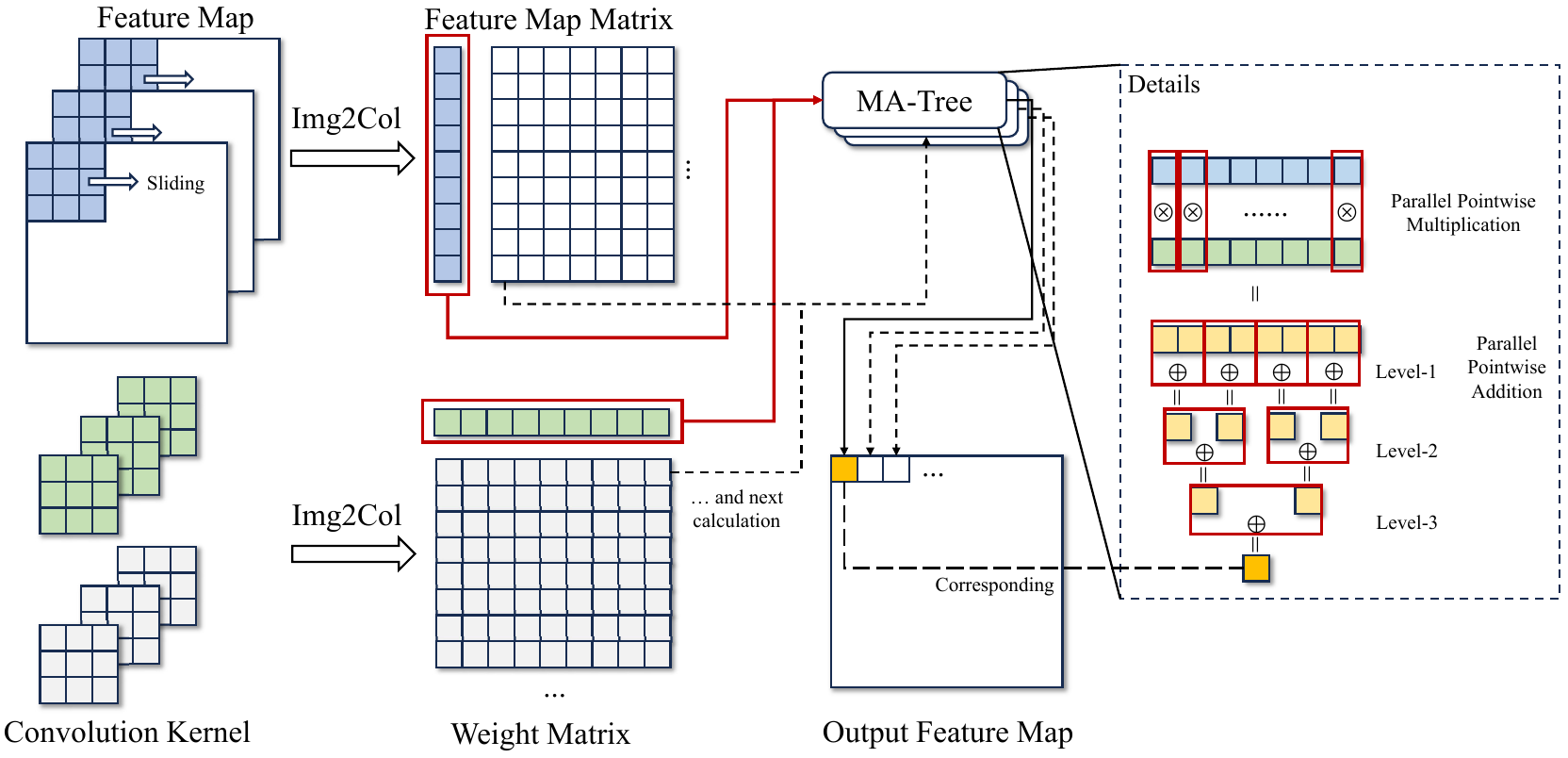}
    \vspace{-0.1in}
    \caption{
    This is an overall illustration of Img2Col and Multiplication-Add Tree. The Img2Col method converts convolutional computation into matrix multiplication by rearranging the data in the feature map and convolution kernel. Then, these matrices will be fed into arrays consisting of Multiplication-Addition Trees. After a step-by-step and level-by-level calculation, we obtain elements of the result and combine them into an output matrix.
    }
    \label{fig:img2col}
\end{figure}

Much of the literature shows that the primitive way of computing convolution requires frequent jumps to access the data, which is a cache-unfriendly way of accessing the data, and therefore, it is difficult to fully exploit the processor power by computing the convolution directly by definition. Img2Col is a method of reordering the raw data of a convolution operator to make it suitable for computation. Demonstrated in the left part of Fig.~\ref{fig:img2col}, this method converts the convolution operation into a matrix multiplication operation, thus concentrating the originally dispersed convolution kernel into a local region, which can increase the probability of local access during computation, and at the same time, it also reduces a kind of operation that should be designed for complex computation mechanism to the same computation as the fully connected one, which unifies the computation method, reduces the cost of design, and facilitates the unification of perception of its operation cycle and performance.\\
To perform the calculation, Img2Col first expands the convolution kernel and feature map into matrices and multiplies them together to produce the output, with the convolution kernel matrix on the left and the feature map matrix on the right. For the convolution kernel, each of its output channels is treated as a row of the convolution kernel matrix, and the data in that output channel is filled into that row in the order in which it was computed. This results in a convolution kernel matrix of size $(k_o) \times (k_i,\ k_h,\ k_w )$. For the feature map, the data multiplied each time by the convolution kernel counterpart is arranged in order as a column of the feature map matrix, with different columns for the data at different locations covered multiple times by the convolution kernel. This gives the size of the feature map matrix as $(in_c,\ k_h,\ k_w ) \times (out_h,\ out_w)$, under known condition $k_i=in_c$. Multiplying the two gives the output matrix with dimensions $(k_o ) \times (out_h,\ out_w )$, whose data is arranged in the same order as the input feature maps and can be used to compute the next layer of operators in the neural network without extra rearranging.\\
\\
On the whole, this rearrangement method is mainly performed from two perspectives: for the input feature map, it flattens and splices the data of different channels at the same sliding window position; for the convolution kernel, it treats different channels in the same convolution kernel in the same way and finally obtains the matrix obtained by splicing the columns from the window transformations. In this way, the data scattered in space becomes continuous during computation, making it easy to hit the cache and convenient for FPGA transfer and computation logic.\\

\subsubsection{Multiplication-Addition Tree}
\label{matree_sec}
Since we have converted the convolution operation into a matrix operation in Img2Col, as shown in the right part of Fig.~\ref{fig:img2col}, we can use only one computational mechanism to complete the matrix multiplication operation. Here, we use the Multiplication-Addition Tree to complete the matrix operation, whose operation mechanism regards vector dot product as the basic calculation step. Each time the calculation of the input is a pair of two vectors of the same length, that is, the matrix operation of the two input matrices in the selection of columns or rows of data. The multiplication tree will group the elements at the corresponding positions of the two inputs into two-by-two groups in a front-to-back order, and then multiply each group internally to get the result of the group. After the point multiplication is the continuous addition with the accumulation of the intermediate results, where there is no dependence on each other, so we can continue to divide and calculate them in parallel with each other, forming a binary tree. Besides being able to parallelize the elements of the same step of the operation, such a computational mechanism can also be employed to add the appropriate flow processing logic, so that manage the data stream continuously.\\
\\
\subsubsection{Sub Matrix Slicing}
\label{submat_sec}
We have already looked at how to convert and compute convolutions and other matrix multiplication operations, now we look at optimizing matrix multiplication. Consider a matrix multiplication between two matrices of side length $L$, whose sliced blocks are of side length $M$. Let $L$ be an integer multiple $N$ of $M$. Each matrix can be sliced into $N^2$ sub-matrix blocks of side length M and requires $N^3$ sub-matrix multiplications and the corresponding accumulations to obtain the final result. In these $N^3$ sub-matrix multiplications, two input matrices of size $M^2$ and one result matrix must be transmitted each time, i.e. the total amount of data to be transmitted is:
\begin{equation} \label{xxx}
    C=3×M^2×N^3=\frac{3L^3}{M}
\end{equation}
\begin{algorithm}[t]
\caption{Split Matrix} 
\label{al1}
\begin{multicols}{2}
\hspace*{0.02in} {\bf Input:} 
Length of the side of each weight or each feature map matrix $\{L_1, L_2, ..., L_n\}$\\
\hspace*{0.02in} {\bf Output:} 
The tile side length $\{T_1, T_2, \cdots, T_n\}$ of each layer
\begin{algorithmic}[1]
\State $L_{\text{max}}\rightarrow \max\{L_1, L_2,\cdots, L_n\}$ 
\State $L_{\text{min}}\rightarrow \min\{L_1, L_2,\cdots, L_n\}$ 
\For{$i=1,2,...,n$} 
  \If{$L_i \le L_{\text{max}}$} 
    \If{$L_i$ is power of 2}
        \State $T_i\leftarrow L_{\text{max}}$
    \Else \If{$L_i<L_{\text{min}}$}
            \State $T_i\leftarrow L_{\text{min}}$
        \Else
            \State $n\leftarrow\text{floor}(\log_2 L_i)$
            \If {$L_i > \dfrac{2^n+2^{n+1}}{2}$}
                \State $T_i\leftarrow 2^{n+1}$
            \Else
                \State $T_i\leftarrow 2^{n}$
            \EndIf
        \EndIf
    \EndIf
  \Else
    \State $n\leftarrow L_i \mod L_{\text{max}}$
    \If {$n > \dfrac{L_{\text{max}}}{2}$}
        \State $T_i\leftarrow L_{\text{max}}$
    \Else
        \State $T_i\leftarrow \dfrac{L_{\text{max}}}{2}$
    \EndIf
  \EndIf
\EndFor
\State \Return $\{T_1, T_2, \cdots, T_i\}$
\end{algorithmic}
\end{multicols}
\end{algorithm}
It can be seen that the total amount of data transferred $C$ decreases monotonically with the side length $M$ of the sliced submatrix block. In FPGA acceleration of neural networks, there is a trade-off between reducing the total amount of data transferred and reducing inefficient computation. On the one hand, on-chip and off-chip data transfer is often a performance bottleneck, so the total amount of data transferred must be minimized, as shown in Equation \ref{xxx}, and the size of the sliced submatrix block must be increased. On the other hand, since the matrix may need to be extended to align the data during slicing to accommodate the design logic of the FPGA if the original matrix edge length is not a power of 2, this may lead to an increase in the amount of data to be transferred, especially the invalid data, so the length of the extension must be reduced and the block size of the sliced sub-matrix urgently reduced. In addition, to facilitate the computation of small matrices, it is necessary to restrict the individual matrices to be able to fill at least one row of the input buffer, i.e. the minimum side length of the matrix $L_{\text{min}} \ge\sqrt{L_{\text{max}}}$ must be satisfied. According to the above principle, the original matrix is chunked: the smallest of the 4 edges of the original matrix is taken as the edge length to be sliced. If the matrix edge length is a power of 2 and does not exceed $L_{\text{max}}$, then this edge length is taken directly as the slicing length; otherwise, if the matrix edge length is less than $L_{\text{min}}$, then $L_{\text{min}}$ is taken as the slicing length; otherwise if it is less than 64, it is extended to the next power of 2. If the matrix edge length is greater than $L_{\text{max}}$, then $L_{\text{max}}$ is taken as the slicing length; otherwise, it is extended to the next power of 2; otherwise extend it to the next power of 2; if the matrix edge length is greater than $L_{\text{max}}$, then take the remainder of the matrix edge length at $L_{\text{max}}$. If the remainder is greater than $L_{\text{max}} /2$, then make the slicing length $L_{\text{max}}$, otherwise make the slicing length $L_{\text{max}} /2$. Using the matrix slicing Algorithm~\ref{al1}, the original matrix can be sliced into appropriate sizes so that the size of each of these matrices does not exceed the capacity of the on-chip memory.\\
\\
\subsubsection{BRAM Merges and Allocation}
\label{bram_sec}
Finally, we need to consider the details of implementing matrix multiplication on a particular model of FPGA. The most important of these is the organization of the data. The ZYNQ platform provides a BRAM structure, which is more suitable for generating memory units for centralized processing than distributed RAM generated using LUTs. Normally BRAM is 72 bits wide, and we use only 64 bits of this as an 8-byte unit to facilitate the alignment process and simplify the logic. Obviously, this width is not enough to handle a large number of large bit-width data, so we usually splice it to get a large bit-width memory space, which is convenient for data transfer, caching and positioning; at the same time, we also need to pay attention to the limit of the amount of BRAM, which cannot exceed the limit that the FPGA can provide, so we use the following Algorithm~\ref{al2} to limit the amount of BRAM usage as a simple segmentation example. The idea here is to allocate BRAM by controlling the input to limit space one column at a time, and the output is cached on the chip as much as possible for accumulation, so the output consumes more BRAM relatively to reduce repeated transmission.

\begin{algorithm}[t]
\caption{BRAM Analyses} 
\label{al2}
\hspace*{0.02in} {\bf Input:} 
BRAM limits $B_{\text{max}}$\\
\hspace*{0.02in} {\bf Output:} 
BRAM allocation for weight, feature map, and output buffer $B_w, B_f, B_o$ and the side length $L_{\text{max}}$ of the maximum matrix to be multiplied.
\begin{algorithmic}[1]
\State let $L_{\text{max}}$ be a power of 2, for example $L_{\text{max}} \leftarrow 8$
\While{true} 
  \State $B_w\leftarrow L_{\text{max}} \times \text{BRAM\_coe\_for\_weight}$       // Depend on the organization of BRAM for weight storage 
  \State $B_f\leftarrow L_{\text{max}} \times \text{BRAM\_coe\_for\_feature\_map}$       // Depend on the organization of BRAM for feature map storage 
  \State $B_o\leftarrow L_{\text{max}} \times \text{BRAM\_coe\_for\_output}$       // Depend on the organization of BRAM for output storage 
  \If{$B_w+B_f+B_o\le B_{\text{max}}$} 
    \State $L_{\text{max}}\leftarrow 2\times L_{\text{max}}$
  \Else
    \State break
  \EndIf
\EndWhile
\State $L_{\text{max}}\leftarrow \dfrac{L_{\text{max}}}{2}$
\State $B_w\leftarrow L_{\text{max}} \times \text{BRAM\_coe\_for\_weight}$
\State $B_f\leftarrow L_{\text{max}} \times \text{BRAM\_coe\_for\_feature\_map}$
\State $B_o\leftarrow L_{\text{max}} \times \text{BRAM\_coe\_for\_output}$
\State \Return $B_w, B_f, B_o, L_{\text{max}}$
\end{algorithmic}
\end{algorithm}


\newpage

\end{document}